\documentclass[10pt,twocolumn,letterpaper]{article}

\usepackage[pagenumbers]{cvpr} 

\usepackage{graphicx}

\newcommand{\Fref}[1]{Figure~\ref{#1}}
\newcommand{\Eref}[1]{(\ref{#1})}

\newcommand{\Tref}[1]{Table~\ref{#1}}

\usepackage{multirow} 
\usepackage{amsfonts}
\usepackage{booktabs}       
\usepackage{amsmath}
\usepackage{algorithm}
\usepackage{algpseudocode}

\usepackage{xcolor}

\definecolor{cvprblue}{rgb}{0.21,0.49,0.74}
\usepackage[pagebackref,breaklinks,colorlinks,allcolors=cvprblue]{hyperref}

\title{MultiDreamer3D: Multi-concept 3D Customization with Concept-Aware Diffusion Guidance}

\author{
    Wooseok Song\textsuperscript{1}, Seunggyu Chang\textsuperscript{2}, and Jaejun Yoo\textsuperscript{1}\\ \\
    \textsuperscript{1}Ulsan National Institute of Science and Technology (UNIST) \\ 
    \textsuperscript{2}NAVER Cloud
}

\begin{document}

\twocolumn[{%
\renewcommand\twocolumn[1][]{#1}%
\maketitle
\begin{center}
    \centering
    \captionsetup{type=figure}
    \includegraphics[width=1.0\linewidth]{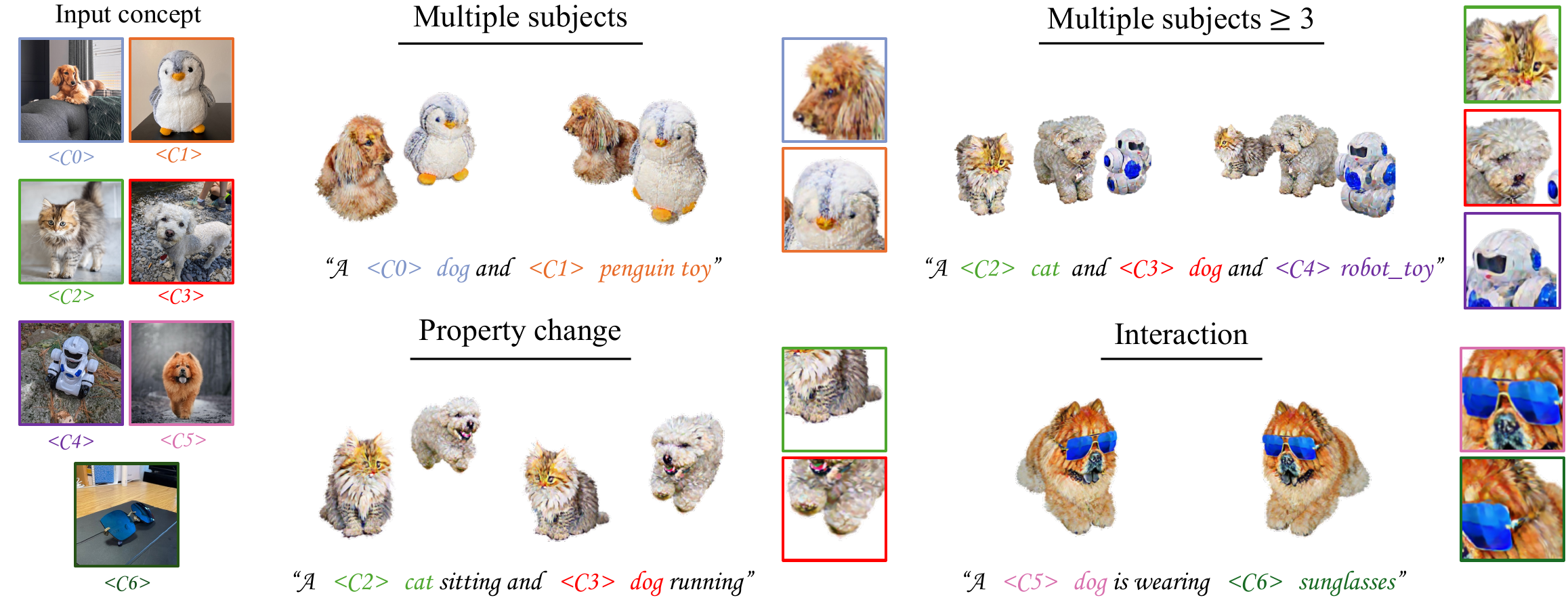} 
    \caption{Multi-concept 3D customization with MultiDreamer3D. MultiDreamer3D can generate 3D content incorporating multiple input concepts in three cases: 1) multiple subjects, 2) property change, and 3) interaction.}
    \label{figure1}
\end{center}%
}]

\begin{abstract}
While single-concept customization has been studied in 3D, multi-concept customization remains largely unexplored. To address this, we propose MultiDreamer3D that can generate coherent multi-concept 3D content in a divide-and-conquer manner. First, we generate 3D bounding boxes using an LLM-based layout controller. Next, a selective point cloud generator creates coarse point clouds for each concept. These point clouds are placed in the 3D bounding boxes and initialized into 3D Gaussian Splatting with concept labels, enabling precise identification of concept attributions in 2D projections. Finally, we refine 3D Gaussians via concept-aware interval score matching, guided by concept-aware diffusion. Our experimental results show that MultiDreamer3D not only ensures object presence and preserves the distinct identities of each concept but also successfully handles complex cases such as property change or interaction. To the best of our knowledge, we are the first to address the multi-concept customization in 3D.
\end{abstract}    
\section{Introduction}
\label{sec:intro}

Recent advancements in text-to-3D methods~\cite{poole2022dreamfusion,liang2023luciddreamer} have significantly progressed the generation of 3D models~\cite{mildenhall2021nerf,kerbl20233d} from text prompts. The main idea is to optimize 3D models by distilling the score of text-to-image diffusion model~\cite{rombach2022high,gal2022image} using score distillation sampling (SDS).  
The SDS enables the generation of both general objects and personalized subjects or concepts, such as ``\textit{one's dog}'' or ``\textit{unique sunglasses}'' with personalized diffusion models~\cite{ruiz2023dreambooth,gal2022image}. 
However, the existing literature predominantly focuses on customizing a single-concept 3D model, thereby constraining its application in more diverse and complex scenarios.

\begin{figure}[!t]
\centering
\includegraphics[width=0.9\columnwidth]{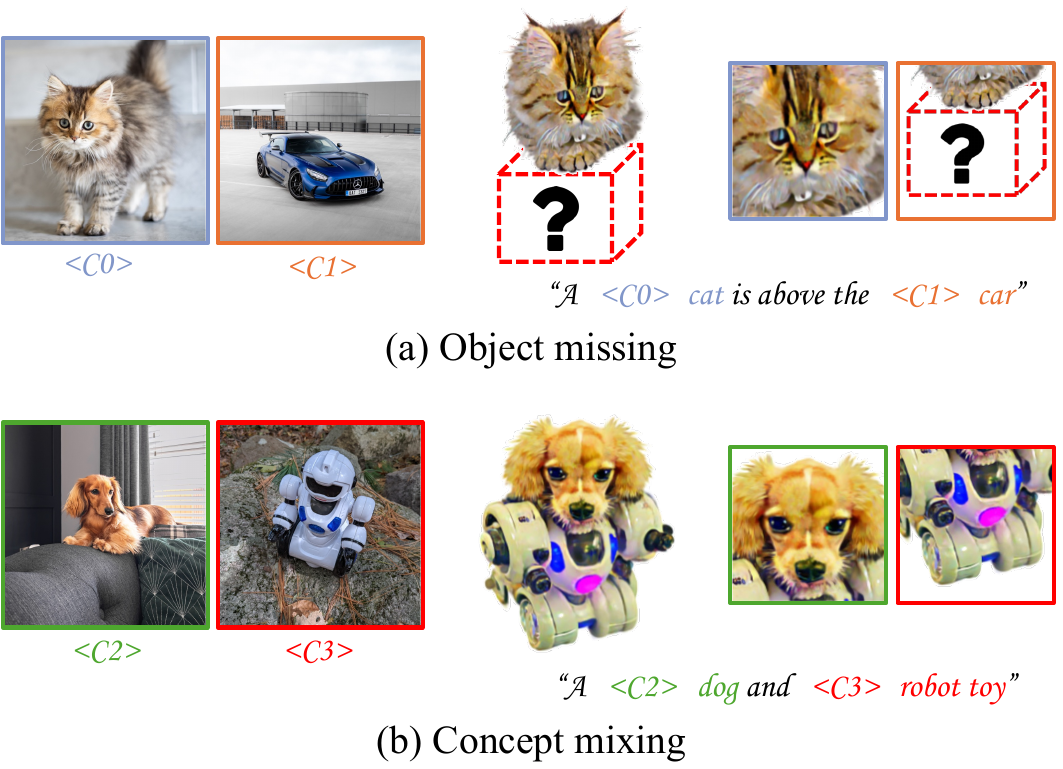} 
    \caption{Challenges in multi-concept 3D customization. The 3D content is produced using multi-concept 2D diffusion models using the SDS-based method~\cite{liang2023luciddreamer}. (a) The ``\textit{\texttt{C1} car}'' is missing which leads to poor layout context. (b) The dog's head is combined with a robot toy's body, which we call a concept-mixing problem.}
\label{fig:figure2_problem}
\end{figure}

In this study, we tackle multi-concept text-to-3D customization, aiming to produce a 3D model that includes multiple user-defined concepts. For example, consider the 3D model generated from the text prompt: ``\textit{A \texttt{C0} dog is wearing \texttt{C1} sunglasses}.'' where \textit{\texttt{C0}} and \textit{\texttt{C1}} represent user-specific concepts such as their ``\textit{one's dog}'' or ``\textit{unique sunglasses}''. Achieving high-quality multi-concept 3D models entails overcoming two main challenges: object missing and concept-mixing problems, as illustrated in~\Fref{fig:figure2_problem}. First, current text-to-3D methods~\cite{poole2022dreamfusion,liang2023luciddreamer} struggle to generate 3D content that accurately represents multiple objects described in a given textual description. This issue arises primarily due to the limitations inherent in text-to-image diffusion models~\cite{rombach2022high,saharia2022photorealistic}, which not only face challenges in generating multiple objects in 2D but also often suffer from poor layout context, leading to missing or incorrectly positioned objects. Second, naively adapting multi-concept 2D diffusion model~\cite{mcmahan2017communication,gu2024mix} to optimize 3D model using SDS struggles with the concept-mixing problem, where distinct concept identities are blended or lost. This issue arises from two main factors: the inherent instability of SDS, and the difficulty in managing multiple concepts within a single 2D diffusion model. When these two components are combined, the resulting 3D model often fails to accurately preserve and distinguish between the multiple user-defined concepts.

To address these challenges, we introduce MultiDreamer3D, a method designed to preserve the individual identities of each concept within a coherent layout context in 3D. The MultiDreamer3D operates in two main stages, utilizing two primary modules: the 3D Layout Generator (LG) and Concept-aware Diffusion Guidance (CDG). In the first stage, LG addresses the object missing by incorporating a large language model (LLM)~\cite{achiam2023gpt} based 3D layout controller and a selective concept point cloud generator. Specifically, we obtain 3D bounding boxes by querying text prompts to the 3D layout controller, ensuring the presence of objects and coherent layout context. Subsequently, the selective concept point cloud generator generates individual coarse point clouds for each concept, referred to as concept point clouds, and positions them within the 3D bounding boxes. In the second stage, CDG addresses the concept-mixing problem by updating the 3D Gaussian with the concept-aware diffusion score. Specifically, 3D Gaussians are initialized with the concept point clouds and explicit concept labels, and updated with the proposed concept-aware interval score matching (CISM) loss. This approach ensures that each concept maintains its distinct identity without blending or loss during 3D model optimization. As illustrated in~\Fref{figure1}, our method can generate 3D models with multiple concepts. To demonstrate the effectiveness of MultiDreamer3D, we construct and evaluate three cases of multi-concept 3D content generation: 1) multiple subjects, 2) property change, and 3) interaction. These cases illustrate how MultiDreamer3D effectively maintains the distinct identities of multiple concepts while ensuring object presence and a coherent layout, even in cases involving complex interactions within a 3D space. Our contributions can be summarized as follows:
\begin{itemize}
    \item To the best of our knowledge, we are the first to address multi-concept 3D customization. 
    \item We introduce a 3D Layout Generator (LG) that generates 3D bounding boxes and individual concept point clouds, addressing the object-missing problem.
    \item We propose Concept-aware Diffusion Guidance (CDG) that updates 3D Gaussians based on concept-aware diffusion score, addressing the concept-mixing problem.
    \item Our experimental results demonstrate the effectiveness of our method, showcasing its ability to maintain distinct concept identities of multiple concepts within a coherent layout context in 3D. 
\end{itemize}

\begin{figure*}[t]
\centering
\includegraphics[width=1.0\textwidth]{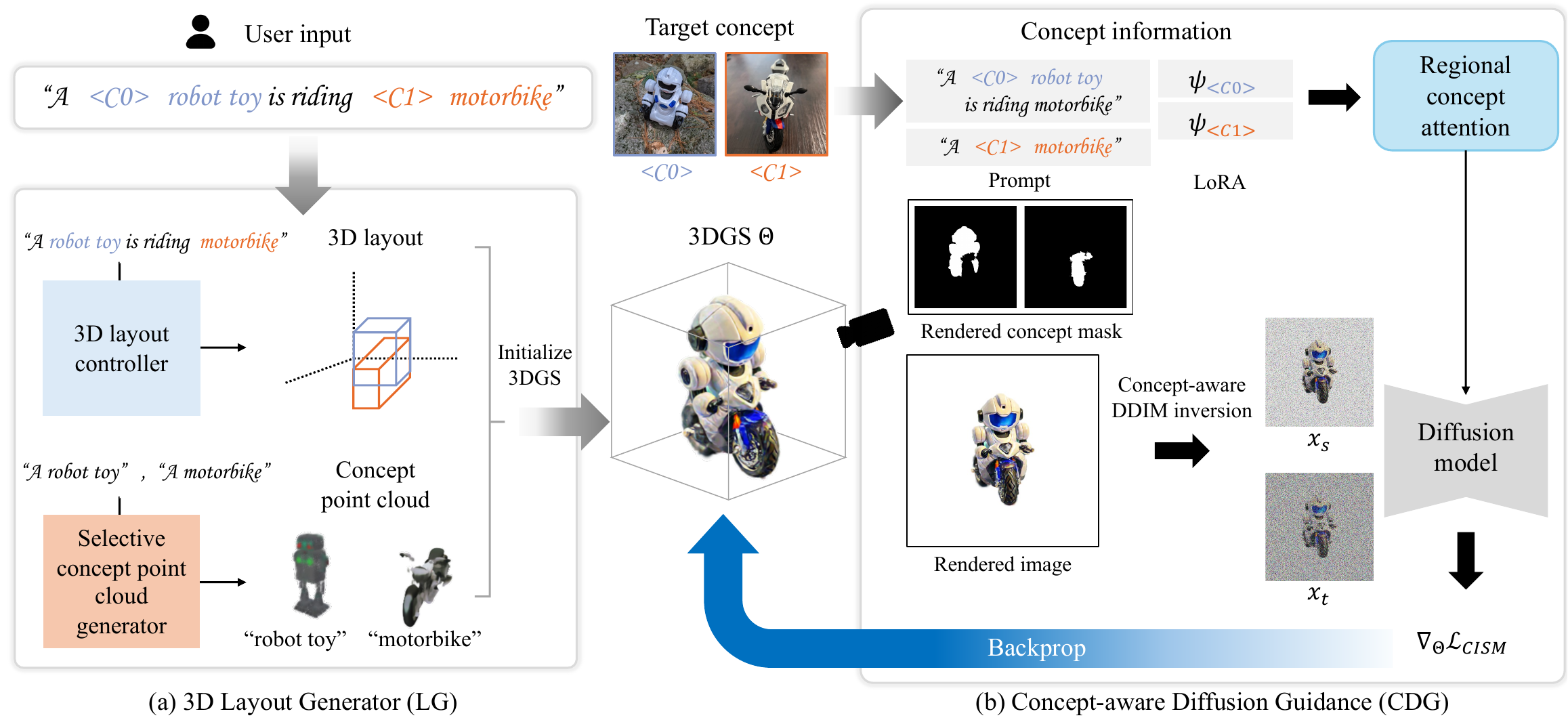} 
\caption{Overall pipeline of MultiDreamer3D. (a) The 3D layout controller produces 3D bounding boxes given text descriptions. Subsequently, the selective concept point cloud generator outputs coarse concept point clouds and positions within the 3D bounding boxes. (b) The images and concept masks are rendered from 3D Gaussian Splatting (3DGS) $\Theta$ and updated with concept-aware interval score matching (CISM) loss, facilitated by regional concept attention (RCA).}
\label{fig:figure_pipeline}
\end{figure*}

\section{Background}
\subsection{3D Gaussian Splatting}
3D Gaussian Splatting (3DGS)~\cite{kerbl20233d} has emerged as a leading explicit 3D representation for novel view synthesis.
3DGS is composed of updatable anisotropic 3D Gaussians denoted as $\Theta=\{\mathbf{\mu},\Sigma,\mathbf{\sigma},\mathbf{c} \}$. Here, $\mathbf{\mu}\in\mathbb{R}^3$ represents the position, $\Sigma\in\mathbb{R}^{3\times3}$ is the 3D covariance, $\mathbf{\sigma}\in\mathbb{R}$ denotes the opacity, and $\mathbf{c}\in\mathbb{R}^s$ represents the color, where $s$ indicates the degree of spherical harmonics (SH). The 3D Gaussian is formulated as follows:
\begin{equation}
    G(\mathbf{x})=e^{-{1\over 2}\mathbf{x}^T\Sigma^{-1}\mathbf{x}}.
\end{equation}

3DGS uses a neural point-based rendering technique for pixel color computation, which involves blending $\mathcal{N}$-ordered overlapping  points:
\begin{equation}
    C=\sum_{i\in \mathcal{N}}{\mathbf{c}_i\alpha_i \prod_{j=1}^{i-1}(1-\alpha_j)}. 
\end{equation}
Here, $\mathbf{c}_i$ refers to the per-point color, and $\alpha_i$ is computed based on the per-point opacity $\sigma_i$ and the 2D projection of the 3D covariance $\Sigma$.

\subsection{Lifting 2D Diffusion Model to 3D}
Score distillation sampling (SDS)~\cite{poole2022dreamfusion} has become a promising method for text-to-3D generation. This technique cleverly adapts the text-to-image diffusion model to optimize 3D models, such as NeRF~\cite{mildenhall2021nerf} or 3DGS~\cite{kerbl20233d}. 
Recently, LucidDreamer~\cite{liang2023luciddreamer} proposed Interval Score Matching (ISM), which aims to improve 3D generation quality by updating $\Theta$ with multi-step noise prediction. The process begins with the prediction of noise $\epsilon_{\phi}(\mathbf{x}_s,\emptyset,s)$ at the diffusion timestep $s=t-\delta_T$. Here, $\delta_T$ indicates the step size of the Denoising Diffusion Implicit Model (DDIM)~\cite{song2020denoising} inversion, and $\emptyset$ denotes null text prompt. Following this, $\textbf{x}_t$ is derived through the DDIM inversion process. The gradient of ISM is calculated as follows:  
\begin{equation}\label{eqn: ism}
      \resizebox{.9\hsize}{!}{$\nabla_{\Theta}\mathcal{L}_{ISM}(\phi,\textbf{x})=\mathbb{E}_{t,\epsilon}\bigl[w(t)(\underbrace{{\epsilon}_\phi(\mathbf{x}_t;y,t)-{\epsilon}_\phi(\mathbf{x}_s;\emptyset,s)}_\textnormal{ISM update direction}){\partial \textbf{x} \over \partial\Theta}\bigr]$.}
\end{equation}
These methods enable the effective transfer of textual descriptions into precise 3D geometries without the need for expensive 3D supervision. 

\section{Method}
The overall pipeline of our method is illustrated in \Fref{fig:figure_pipeline}. Our method consists of two stages, utilizing two primary modules: 1) 3D Layout Generator (LG) and 2) Concept-aware Diffusion Guidance (CDG). In the first stage, the LG generates 3D bounding boxes with a 3D layout controller to specify individual concept objects, considering the layout context. Subsequently, the LG generates and selects point clouds for each concept, termed concept point clouds, with a selective concept point cloud generator that acquires their coarse geometry. These concept point clouds are then positioned within their respective 3D bounding boxes. In the second stage, we initialize a 3DGS with the concept point clouds and assign concept labels to identify the concepts of each 3D Gaussian. The 3D Gaussians are then updated using CDG, specifically through a concept-aware interval score matching (CISM) loss that incorporates regional concept attention (RCA), designed to preserve the distinct identities of the concepts throughout the process.

\subsection{3D Layout Generator}
\label{sec:LC}

\label{sec:LLMlayout}
\textbf{3D Layout Controller.} To produce multi-concept 3D content of high quality, it is essential to ensure both the presence of objects and layout context based on textual descriptions. To address this, we propose a 3D layout controller that leverages Large Language Models~\cite{achiam2023gpt}, which generates 3D bounding boxes for individual concepts based on text prompts. We create examples for three cases (multiple subjects, property change, and interaction) to serve as samples for in-context learning. The 3D layout controller then uses in-context examples with instruction to output the parameter of 3D bounding boxes $Bbox_{i}=[X_i,Y_i,Z_i,W_i,D_i,H_i]$ for each concept in global coordinate system. Then, we derive scale $s_i$ and translation $ t_i$ to position $i$-th concept objects into 3D bounding boxes:
\begin{equation}
    \small {s_i={\min({W_i \over W},{H_i \over H})}, t_i=\left[{X_i+{W_i \over 2}},{Y_i+{D_i \over 2}},{Z_i+{H_i \over 2}}\right]}.
\end{equation}
$W$ and $H$ denote the maximum width and height, ($X_i, Y_i$, $Z_i$) are the coordinates of the lowest left corner, and ($W_i, D_i$, $H_i$) represent the width, depth, and height of the bounding box for the $i$-th concept.

\medskip
\noindent\textbf{Selective Concept Point Cloud Generator.} The goal of concept point cloud generation is to acquire the coarse geometry of individual concepts. To achieve this, we employ Shap-E~\cite{jun2023shap} to generate initial concept point clouds based on text prompts. These prompts can include either simple concept class tokens or brief descriptions, such as ``\textit{a dog}'' or ``\textit{a jumping dog}''. Shap-E then generates implicit neural representation (INR) weights corresponding to these text prompts. Following this, vertices of the voxel grid are queried through the INR to obtain colors and signed distance function values, which are subsequently used to construct the concept point clouds. In our method, we manually input text prompt to Shap-E for each concept point clouds.

However, Shap-E often generates point clouds with distorted geometry. To mitigate this issue, we introduce a point cloud selector that utilizes a vision language model (VLM)~\cite{achiam2023gpt} to ensure reliable 3D geometry. Our selection module begins by generating multiple candidate point clouds from a single text prompt using Shap-E. The point cloud selector then evaluates these candidates by analyzing renderings from fixed viewpoints, selecting the point cloud that best matches the text prompt. 
The selected point cloud $\textbf{pcd}_i$ is then positioned within the 3D bounding box in global coordinate system:
\begin{equation}
    {
        \textbf{pcd}_{global}={s}_i\times\textbf{pcd}_i+{t}_i.
    }
\end{equation}
Here, $s_i$ and $t_i$ denote scale and translation of $i$-th concept.

\begin{figure*}[!t]
\centering
\includegraphics[width=\textwidth]{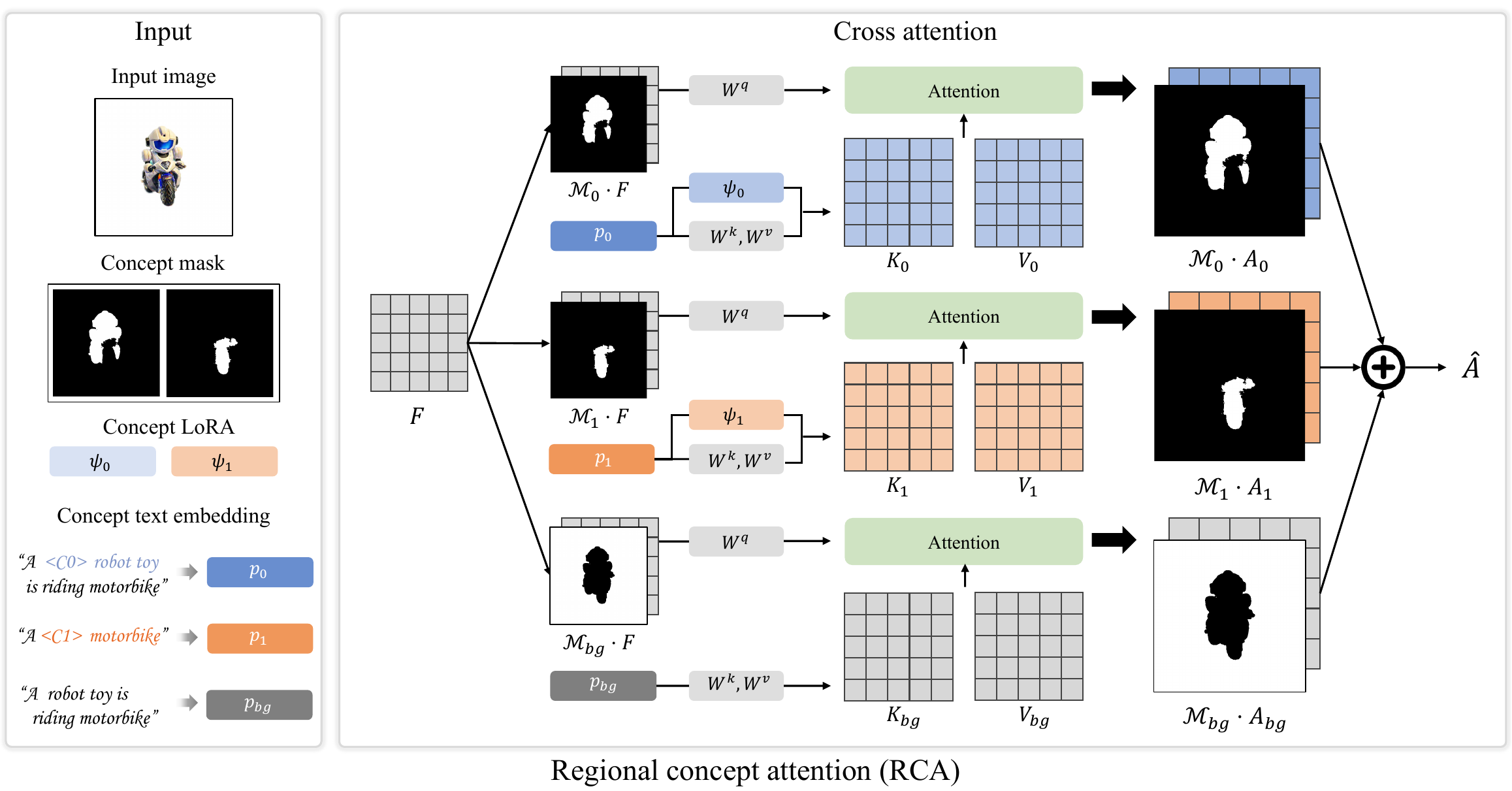} 
\caption{The Regional Concept Attention (RCA) modulates the cross-attention layer in the diffusion model. Individual concept query vectors are computed with image features and each concept masks. Subsequently, key and value vectors for each concept are derived using concept-specific LoRAs and prompts. Then concept-specific attention features are computed with each query, key, and value. The final cross-attention features are aggregated with masked concept-specific attention features.}
\label{fig:RCA}
\end{figure*}

\subsection{Concept-aware Diffusion Guidance}
\label{sec:CISM}
\textbf{3DGS Initialization with Concept Labeling.} After concept point clouds are generated and positioned, they are initialized into 3D Gaussians. However, initializing 3D Gaussian without embedding concept information cannot give precise feedback for individual concepts. To address this, we propose concept labeling by incorporating a $k$-class one-hot encoded concept label $\mathbf{m}\in \mathbb{R}^k$ into each 3D Gaussian, represented as $\Theta_i=\{\mathbf{\mu}_i,\Sigma_i,\mathbf{\sigma}_i,\mathbf{c}_i, \mathbf{m}_i \}$. This setup enables the rendering of a 2D binary concept mask $\mathcal{M}\in\mathcal{R}^{k\times h \times w}$, facilitating precise concept-specific feedback for each Gaussian. The rendering process of concept rendering $M$ follows:
\begin{equation}
    M=\sum_{i\in \mathcal{N}}{\mathbf{m}_i\alpha_i \prod_{j=1}^{i-1}(1-\alpha_j)}.
\end{equation}
Here, $\mathbf{m}$ denotes the concept label. The concept rendering $M_k\in \mathbb{R}^{1\times h \times w}$ represents the contribution of the $k$-th concept to the projected 2D pixel within the range $[0,1]$. However, this includes low-concept contributions that are noisy. To minimize such noisy contributions, we apply a threshold factor $\tau$ to the concept rendering $M$, producing a binary concept mask $\mathcal{M}\in \mathbb{R}^{ k\times h\times w}$.

\medskip
\noindent\textbf{Regional Concept Attention.} Updating the concept 3D Gaussians with concept-specific feedback is essential to prevent concept mixing. To achieve this, we introduce the Regional Concept Attention (RCA) module as shown in~\Fref{fig:RCA}. The RCA modulates the cross-attention map in a text-to-image diffusion model~\cite{rombach2022high} by incorporating individual concept information. This enables unified noise prediction while preserving the distinct identities of each concept. The noise prediction process is as follows.

\begin{figure*}[!t]
\centering
\includegraphics[width=\textwidth]{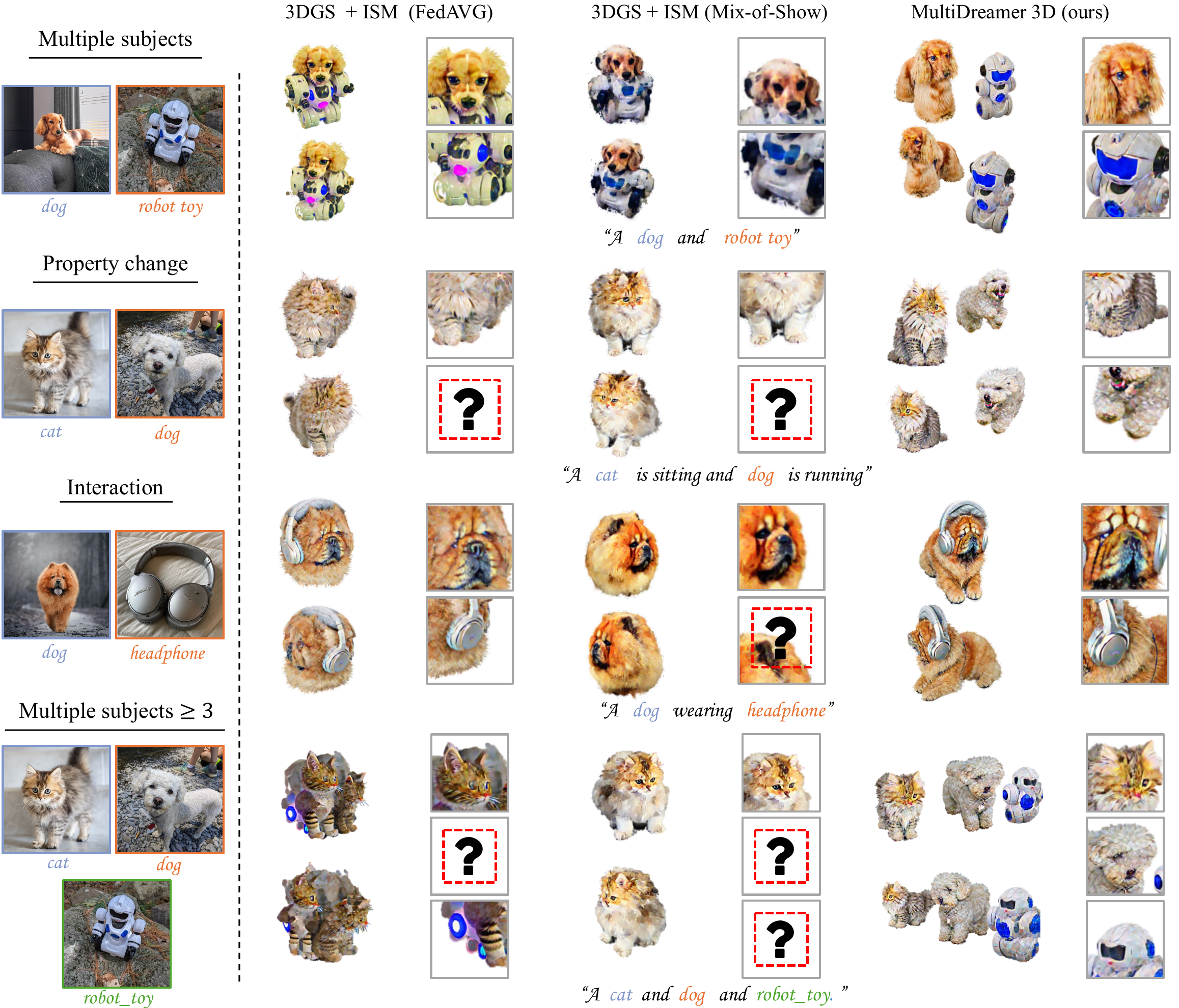} 
\caption{Qualitative results. We compare our method with other baselines in three cases, mulitple subjects, property change, and interaction. The red dashed line indicates the objects mentioned in the text prompt that are missing.}
\label{fig:qualitative results}
\end{figure*}

First, we observe that text prompts containing multiple concepts often lead to concept-mixing problems, as illustrated in ~\Fref{fig:figure2_problem} (b). To address this issue, we manuallydecompose the text prompts into individual concept prompts. For example, when generating a 3D model from a text prompt \textit{``A \texttt{C0} robot toy is riding \texttt{C1} motorbike''}, we break it down into the following concept prompts:
\begin{align*}
p_{0} &= \textit{``A \texttt{C0} robot toy is riding motorbike''}, \\
p_{1} &= \textit{``A \texttt{C1} motorbike''}, \\
p_{bg} &= \textit{``A robot toy is riding motorbike''}.
\end{align*}

\noindent Next, we modulate the cross-attention layer with the RCA. The RCA inputs concept masks $\mathcal{M}$, concept LoRAs $\psi$, and concept prompts $p$ and outputs an aggregated concept-specific attention feature. The concept-specific query vector is computed:
\begin{equation}
Q_i=W^q\cdot (\mathcal{M}_i\cdot F), Q_{bg}=W^q\cdot (\mathcal{M}_{bg}\cdot F).
\end{equation}
Here, $W^q$ denotes the query projection matrix and $F$ denotes the input image feature. $\mathcal{M}_{i}$ denotes the $i$-th concept mask, while $\mathcal{M}_{bg}=(\mathcal{M}_{0}\cup\mathcal{M}_{1} ... \cup \mathcal{M}_{k})^c$ represents the background mask. This process ensures isolated concept query vectors are used for attention computation. Subsequently, concept-specific keys and values are computed:
\begin{align}\label{eqn:k_v}
&K_i= (W^k+\lambda \cdot \psi_i^k)\cdot p_i, V_i= (W^v+\lambda \cdot \psi_i^v)\cdot p_i\\
&K_{bg}= W^k\cdot p_{bg}, V_{bg}= W^v\cdot p_{bg}.
\end{align}
Here, $W^k$ and $W^v$ denote the key and value projection matrices. The $\psi_i$ and $p_i$ represent the $i$-th concept LoRA and the concept prompt, while $\lambda$ is the LoRA scale. This ensures that individual concept information is encoded into keys and values. Then, concept-specific attention features are computed:
\begin{equation}\label{eqn:attention}
A_i=\text{Softmax}\left({{Q_iK_i^T}\over {\sqrt{d}}} \right)\cdot V_i.
\end{equation}
Here, $A_i$ denotes the concept-specific attention feature. Finally, we aggregate concept-specific attention features: 
\begin{equation}\label{eqn:attention_final}
\hat{A}(\Psi,P,\mathcal{M})=\mathcal{M}_{bg}\cdot A_{bg}+\sum_{i=1}^k\mathcal{M}_i\cdot A_i.
\end{equation}
Here, $\hat{A}$ represents the aggregated attention feature, and $\Psi$, $P$ denote the set of concept LoRAs and text prompts. The $k$ denotes the number of concepts. The noise prediction with our RCA module is represented as $\epsilon_{\phi}(\textbf{x}_t,t,\hat{A}(\Psi,P,\mathcal{M}))$. 

\medskip
\noindent\textbf{Concept-aware Interval Score Matching.} We introduce concept-aware interval score matching (CISM), a method designed to optimize each concept's 3D Gaussians using concept-aware diffusion scores. The process begins by rendering a novel view image $\textbf{x}$ and a concept mask $\mathcal{M}$ from the 3D Gaussian $\Theta$. Let $\textbf{x}_t$ and $\textbf{x}_s$ denote latents at timesteps $t$ and $s$, where $s=t-\delta_T$, that are derived through DDIM inversion \cite{song2020denoising} with null text prompts (i.e. `` ''). However, DDIM inversion using a single weight diffusion model~\cite{rombach2022high} lacks concept-specific knowledge, leading to suboptimal inversion results. To overcome this limitation, we introduce concept-aware DDIM inversion, which adapts the RCA module during the inversion process to incorporate multi-concept knowledge. The proposed concept-aware DDIM inversion is formulated as:
\begin{equation}\label{eqn:concept-aware DDIM}
\mathbf{x}_t=\sqrt{\bar\alpha_t}\hat{\mathbf{x}}_0^s+\sqrt{1-\bar{\alpha}_t}\epsilon_\phi(\mathbf{x}_s,s,\hat{A}(\Psi,\emptyset,\mathcal{M})).
\end{equation}
Here, $\hat{\mathbf{x}}_0^s={{1}\over{\sqrt{\bar{\alpha}_s}}}\mathbf{x}_s-{{\sqrt{1-\bar\alpha_s}}\over{\sqrt{\bar\alpha_s}}}\epsilon_\phi(\mathbf{x}_s,s,\hat{A}(\Psi,\emptyset,\mathcal{M}))$, and $\emptyset$ and $\hat{A}(\cdot)$ denote null text prompts and aggregated concept features using the RCA module, respectively. Technically, the null text prompt is tokenized into a \textit{\texttt{\textless BOS\textgreater}}~token followed by a sequence of \textit{\texttt{\textless EOS\textgreater}}~tokens of maximum token length, which is encoded into a null text embedding via a text encoder. The null text embedding is then processed by the RCA module to produce a unconditional part of the concept-aware diffusion score. This diffusion score is subsequently used to predict $x_s \rightarrow x_t$ with Eq.~\Eref{eqn:concept-aware DDIM}. The CISM loss is then computed using the following equation:
\begin{equation}\label{eqn: cism}
\small{
\begin{split}
\nabla_{\Theta}\mathcal{L}_{CISM} = \mathbb{E}_{t,\epsilon}[w(t)(&{\epsilon}_{\phi}(\mathbf{x}_t;t,\hat{A}(\Psi,P,\mathcal{M})) \\
& - {\epsilon}_{\phi}(\mathbf{x}_s;s,\hat{A}(\Psi,\emptyset,\mathcal{M}))){\partial \mathbf{x} \over \partial\Theta}].
\end{split}}
\end{equation}
Here, $\Psi$ and $\mathcal{M}$ denote concept LoRA and masks, while $P$ and $\emptyset$ denote the set of concept prompts and null prompts, respectively. Using the CISM loss, we can effectively update the 3D Gaussian, ensuring individual concept identities.

\begin{table*}[!t]
\centering
\fontsize{9pt}{11pt}\selectfont
\begin{tabular}{lcc}
    \toprule
    \textbf{Method} & \textbf{Text-align} $\uparrow$ & \textbf{Image-align} $\uparrow$ \\ \midrule
    3DGS + ISM with Mix-of-Show~\cite{gu2024mix} & 0.2024 & N/A \\
    3DGS + ISM with FedAVG~\cite{mcmahan2017communication} & 0.2396 & N/A \\
    LG + ISM with Mix-of-Show~\cite{gu2024mix} & 0.2199 & 0.6081 \\
    LG + ISM with FedAVG~\cite{mcmahan2017communication} & 0.2578 & 0.6338 \\
    MultiDreamer3D (Ours) & \textbf{0.2732} & \textbf{0.6582} \\ \bottomrule
\end{tabular}
\caption{Quantitative results. We assess the text-concept alignment  with 3D models using CLIP scores. Here, ours is LG + CISM. 
}
\label{quantitative_table}
\end{table*}

\section{Experiments}
\subsection{Datasets.} We selectively choose real concept image data from the Custom Diffusion~\cite{kumari2023multi} and DreamBooth~\cite{ruiz2023dreambooth} datasets, which contain 13 unique objects (three wearables and 10 unique objects). This selection is made to explore three specific cases: 1) multiple subjects, 2) property change, and 3) interaction. First, the multiple subjects case involves generating 3D models that incorporate several distinct objects simultaneously. Second, the property change case focuses on subjects with altered attributes, such as different poses (e.g., ``\textit{jumping}'' or ``\textit{sitting}''). Third, the interaction case examines where multiple subjects interact in complex ways, such as one subject ``\textit{wearing}'' another. These cases evaluate MultiDreamer3D's ability to both preserve concept identity and maintain the presence of objects while handling complex cases such as property changes or interactions. To comprehensively address these cases, we craft and utilize 47 text prompts specifically designed to cover these three cases.

\subsection{Baseline Methods.} In the absence of multi-concept customization method in 3D, we devise a series of baseline methods using existing 2D approaches. The most intuitive and straightforward baseline involves adapting multi-concept 2D diffusion model to train a single 3D model with interval score matching (ISM)~\cite{liang2023luciddreamer}. Here, we establish two baselines: 3DGS + ISM with FedAVG~\cite{mcmahan2017communication} and Mix-of-Show~\cite{gu2024mix}. For the 3DGS, we initialize the 3D Gaussian using a randomly generated sphere. In the FedAVG approach, multiple single-concept DB-LoRA weights~\cite{ruiz2023dreambooth} are merged into a single LoRA weight using a weighted sum. Similarly, in the Mix-of-Show method, multiple ED-LoRA weights~\cite{gu2024mix} are merged using a gradient fusion technique. Both the single-concept DB-LoRA and ED-LoRA models are trained on 13 unique objects before applying these techniques for multi-concept training. Implementation details are in supplementary materials.

\subsection{Evaluation Metrics.} We evaluate both text-3D and image-3D alignments with CLIP~\cite{radford2021learning}. For text-3D alignment, we render 30 evenly spaced views within an azimuth range of $[-45,45]$ degrees to avoid occlusion and compute the average CLIP score between the text prompt and these renders. For image-3D alignment, we decompose each concept 3D Gaussians with our concept labeling, rendering each isolated concept from 120 views spanning $[-180,180]$ degrees, which are compared to the corresponding real concept images to assess alignment fidelity.

\begin{table}[!t]
\centering
\resizebox{\linewidth}{!}{%
\begin{tabular}{lcc}
    \toprule
    \textbf{Method} & \textbf{Text-align} $\uparrow$ & \textbf{Image-align} $\uparrow$ \\ \midrule
    3DGS + ISM with Mix-of-Show & 1.62 & 1.91 \\
    3DGS + ISM with FedAVG & 2.17 & 2.18 \\
    MultiDreamer3D (Ours) & \textbf{4.72} & \textbf{4.66} \\ \bottomrule
\end{tabular}%
}
\caption{User study. Participants rate alignment on a 5-point Likert scale (1 indicating strong disagreement, 5 indicating strong agreement). Here, ours is LG + CISM. }
\label{user_study}
\end{table}

\subsection{Qualitative Results.} In \Fref{fig:qualitative results}, we compare our method with other baseline methods. Both 3DGS + ISM with FedAVG and 3DGS + ISM with Mix-of-Show struggle to preserve individual concept identities, leading to concept mixing and/or object missing. In contrast, our method demonstrates a superior ability to generate 3D content featuring multiple concepts while effectively maintaining both the presence of objects and the distinct identities of each concept in terms of three cases. In multiple subjects, our method preserves concept identities and aligns with text prompts, avoiding the concept mixing seen in other methods. In property changes, our approach maintains concept integrity and enables pose variations, while others often miss objects or cannot achieve pose variations. In interaction, our method performs comparably to 3DGS + ISM (FedAVG) and better than 3DGS + ISM (Mix-of-Show), effectively capturing complex interactions.

\subsection{Quantitative Results.} In~\Tref{quantitative_table}, we evaluate the image-3D and text-3D alignments of generated outputs. Our method achieves the highest text and image alignment scores, which indicates that our method faithfully reflects text descriptions into multi-concept 3D content  while preserving the identities of individual concepts. For image alignment, since other baselines are initialized with a random sphere, isolating the concept 3D Gaussians for these baselines is not feasible. For fair comparison, we utilize our 3D Layout Generator (LG) module to initialize the 3DGS (third and fourth rows of \Tref{quantitative_table}). 

\subsection{User Study.} To demonstrate the effectiveness of our method, we conduct a user study with 32 participants. The study compares 10 3D samples, where participants evaluate three methods based on two criteria: 1) text alignment, assessing how well the 3D model reflects the text prompts, and 2) image alignment, measuring how accurately the 3D model represents real concept images. Participants rate each model on a 5-point Likert scale~\cite{joshi2015likert}, where 1 signifies ``strongly disagree'' and 5 signifies ``strongly agree''. The results are presented in~\Tref{user_study}. Our method achieves the highest human preference for both text and image alignment across all baselines, demonstrating its ability to accurately reflect text prompts and real concept images.

\begin{figure}[!t]
\centering
\includegraphics[width=1.0\columnwidth]{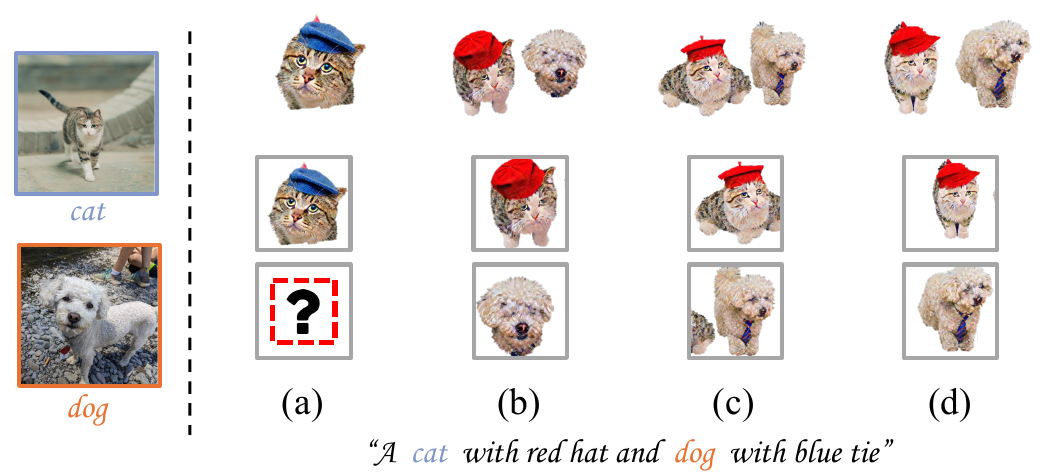} 
    \caption{Ablation study. (a) generated with baseline (3DGS + FedAVG~\cite{mcmahan2017communication}), (b) with proposed 3D Layout Controller + CISM, (c) combined with Shap-E~\cite{jun2023shap}, (d) combined with pointcloud selection.}
\label{fig:figure7}
\end{figure}

\begin{table}[!t]
\centering
\resizebox{\linewidth}{!}{%
\begin{tabular}{lcc}
    \toprule
    \textbf{Components} & \textbf{Text-align} $\uparrow$ & \textbf{Image-align} $\uparrow$ \\ \midrule
    
    Baseline (3DGS + FedAVG) & 0.2396 & N/A \\
    (+) 3D Layout Controller + CISM & 0.2637 & 0.6011 \\
    (+) Shape-E & 0.2720 & 0.6487 \\
    (+) Pointcloud selection (ours) & \textbf{0.2732} & \textbf{0.6582} \\ \bottomrule
    
\end{tabular}%
}
\caption{Ablation study. For the ablation study, we used 3DGS + FedAVG~\cite{mcmahan2017communication} for the baseline.}
\label{ablation}
\end{table}

\subsection{Ablation Study.} In \Fref{fig:figure7} and \Tref{ablation}, we demonstrate the effectiveness of the components in our method. \Fref{fig:figure7} (a) shows the generation of the baseline model (3DGS + FedAVG), which suffers from object missing. \Fref{fig:figure7} (b) presents the generation using our 3D Layout Controller with CISM, which successfully maintains the presence of individual objects. \Fref{fig:figure7} (c) showcases improved geometry in the generated outputs but still suffer from distorted geometry. \Fref{fig:figure7} (d) highlights further enhanced results enabled by the selection module.

\section{Conclusion}
In this paper, we introduced MultiDreamer3D, a method for multi-concept 3D customization that effectively addresses the challenges of object missing and concept mixing. Our 3D Layout Generator facilitates the presence of concept objects and coherent layout context through the use of a 3D layout controller and selective concept point cloud generator. By initializing 3D Gaussian Splatting with explicit concept labeling, we enable clear concept identification. The subsequent update of the 3D Gaussians using Concept-aware Diffusion Guidance ensures the preservation of distinct identities of each concept. Our results showed that MultiDreamer3D is effective across various baselines.

{
    \small
    \bibliographystyle{ieeenat_fullname}
    \bibliography{main}
}


\end{document}